\documentclass[lettersize,journal]{IEEEtran}

\usepackage{amsmath,amsfonts}
\usepackage{algorithmic}
\usepackage{algorithm}
\usepackage{array}
\usepackage[caption=false,font=normalsize,labelfont=sf,textfont=sf]{subfig}
\usepackage{textcomp}
\usepackage{stfloats}
\usepackage{url}
\usepackage{verbatim}
\usepackage{graphicx}
\usepackage{cite}
\usepackage{mathrsfs}
\usepackage{booktabs}
\usepackage{bbm}
\usepackage{bm}
\usepackage{ulem}
\usepackage{bbding}
\usepackage{multirow}
\usepackage[table]{xcolor}

\definecolor{lightyellow}{RGB}{255, 200,150}
\definecolor{lightgray}{RGB}{200,200,200}
\newcommand{\bestcell}{\cellcolor{lightyellow}}
\newcommand{\secondcell}{\cellcolor{lightgray}}

\newcommand{\bx}{\mathbf{x}}
\newcommand{\bz}{\mathbf{z}}
\newcommand{\bc}{\mathbf{c}}
\newcommand{\be}{\bm{\mathbf{\epsilon}}}
\newcommand{\bF}{\mathbf{F}}
\newcommand{\bS}{\mathbf{S}}
\newcommand{\bM}{\mathbf{M}}

\begin{document}

\title{Eliminating Contextual Prior Bias for Semantic Image Editing via Dual-Cycle Diffusion}

\author{Zuopeng Yang, Tianshu Chu, Xin Lin, Erdun Gao, Daqing Liu, Jie Yang, \IEEEmembership{Senior Member, IEEE}, and Chaoyue Wang
\thanks{Manuscript received February 05, 2023; revised April 11, 2023; accepted June 11, 2023. Date of publication June 13, 2022; date of current version June 13, 2023. This research is partly supported by NSFC, China (No: 61876107, U1803261). This article was recommended by Associate Editor L. Jiang.
(Corresponding author: Jie Yang.)

Zuopeng Yang, Tianshu Chu, and Jie Yang are with the Institute of Image Processing and Pattern Recognition, Department of Automation, Shanghai Jiao Tong University, Shanghai 200240, China (e-mail: \{yzpeng, chutianshu, jieyang\}@sjtu.edu.cn). Xin Lin is with the Institute of Artificial Intelligence and Blockchain, Guangzhou University, Guangzhou 510006, China (e-mail: linxj68@gmail.com). Erdun Gao is with the School of Mathematics and Statistics, The University of Melbourne, Melbourne, VIC 3010, Australia (e-mail: erdun.gao@student.unimelb.edu.au). Daqing Liu and Chaoyue Wang are with the JD Explore Academy, Beijing 100176, China (e-mail: liudq.ustc@gmail.com, chaoyue.wang@outlook.com).}}

\markboth{Journal of \LaTeX\ Class Files,~Vol.~14, No.~8, August~2021}%
{Shell \MakeLowercase{\textit{et al.}}: A Sample Article Using IEEEtran.cls for IEEE Journals}

\maketitle

\IEEEpeerreviewmaketitle

\begin{abstract}
The recent success of text-to-image generation diffusion models has also revolutionized semantic image editing, enabling the manipulation of images based on query/target texts. Despite these advancements, a significant challenge lies in the potential introduction of contextual prior bias in pre-trained models during image editing, \textit{e.g.}, making unexpected modifications to inappropriate regions. 
To address this issue, we present a novel approach called Dual-Cycle Diffusion, which generates an unbiased mask to guide image editing. The proposed model incorporates a Bias Elimination Cycle that consists of both a forward path and an inverted path, each featuring a Structural Consistency Cycle to ensure the preservation of image content during the editing process. The forward path utilizes the pre-trained model to produce the edited image, while the inverted path converts the result back to the source image. The unbiased mask is generated by comparing differences between the processed source image and the edited image to ensure that both conform to the same distribution. Our experiments demonstrate the effectiveness of the proposed method, as it significantly improves the D-CLIP score from $0.272$ to $0.283$. The code will be available at \url{https://github.com/JohnDreamer/DualCycleDiffsion}.

\end{abstract}


\section{Introduction}

\begin{figure}[!t]
	\centering \vspace{-0 mm}
	\includegraphics[width=\linewidth]{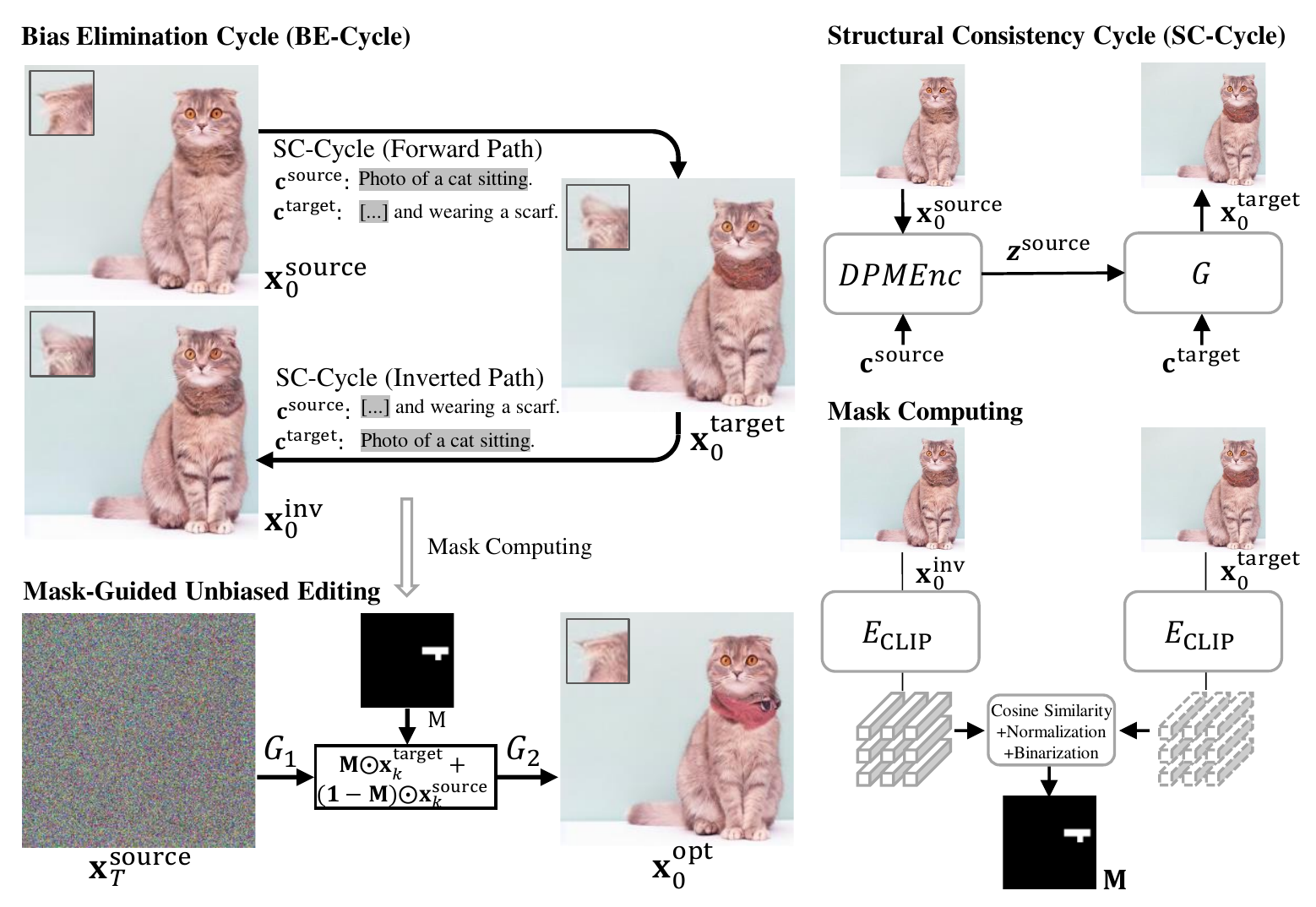}
    \vspace{-7 mm}
    \caption{The overview of the proposed Dual-cycle Diffusion framework for the semantic image editing task. The pipeline and details of the Bias Elimination Cycle (BE-Cycle) and mask-guided unbiased editing are shown on the left. \textbf{The modifications on cat ears}, caused by \textbf{the contextual prior bias derived from the pre-trained model}, are illustrated in the left-top corner of the images. Given a source image, a source text, and a target text, we first leverage BE-Cycle to produce an unbiased mask, which is then used to guide image editing. On the right, the details of the Structural Consistency Cycle (SC-Cycle)~\cite{wu2022unifying} and the procedure of mask computing are shown. $\odot$ is the element-wise product.} 
    \vspace{-4 mm}
	\label{fig:model}
\end{figure}

\IEEEPARstart{S}{emantic} image editing is a critical and demanding problem within the field of image processing, with the objective of modifying an existing image based on a specified textual transformation query. Unlike conventional image editing techniques~\cite{liu2022towards}, such as those utilizing depth~\cite{luo2019disocclusion}, skeleton~\cite{zhang2021lightweight}, edge~\cite{xu2020e2i}, heatmap~\cite{yan2022texture}, or segmentation~\cite{zheng2022semantic,gu2019mask}, the utilization of text~\cite{avrahami2022blended,zhang2021text,li2020manigan,xu2022predict} to guide the editing process offers a more versatile and user-friendly approach to achieving desired modifications to the image. As illustrated in Fig~\ref{fig:model}, given an image of a cat and a query ``\textit{Photo of a cat sitting and wearing a scarf}", the aim is to add a scarf to the cat's neck while maintaining the integrity of the other regions of the image. Thus, this task can be regarded as an extension of text-conditioned image generation, with a precise and interpretable differentiation between regions that require modification and those that should be preserved.

In recent years, there has been substantial progress in text-conditioned generation techniques, with the advent of models such as DALL-E~\cite{ramesh2021zero}, Make-A-Scene~\cite{gafni2022make}, Imagen~\cite{saharia2022photorealistic}, Parti~\cite{yu2022scaling}, and DALL-E2~\cite{ramesh2022hierarchical}.
These models are trained on enormous amounts of data collected from the Internet, resulting in significant improvements in the capacity for both textual semantic understanding and image synthesis.  
Among these models, diffusion-based models have garnered significant attention due to the iterative optimizations from random Gaussian noise to photo-realistic images.

As a pioneering method, SDEdit~\cite{meng2021sdedit} achieves the editing through iterative denoising via a stochastic
differential equation (SDE), which tends to modify the entire image. Prompt2prompt~\cite{hertz2022prompt} localizes the edit by controlling the cross-attention maps to eliminate unexpected modifications only on synthesized images. 
DiffEdit~\cite{couairon2022diffedit} automatically generates a mask by contrasting the predictions of a diffusion model conditioned on different text prompts. However, it has been observed that it fails to produce masks focusing on the regions requiring editing.
Furthermore, to preserve the integrity of the image, CycleDiffusion~\cite{wu2022unifying} presents the DPMEncoder to encode the latent code.

Despite the benefits of utilizing pre-trained models for semantic image editing, a significant challenge arises in the form of prior bias introduced by distribution shifts between the training and target editing images. In this paper, we specifically focus on the contextual prior bias problem, which often leads to spurious correlations between different regions, thus resulting in undesired modifications in areas that were not intended to be altered, but still meet the contextual requirements of the target text. For instance, as exemplified in the top-left corner of Fig~\ref{fig:model}, when attempting to add a scarf to the cat's neck, the pre-trained model fails to maintain the shape of the ears.

To the best of our knowledge, our paper is one of the first studies to eliminate the contextual prior bias in text-guided semantic image editing. We introduce Dual-Cycle Diffusion, a method that effectively eliminates contextual prior biases by generating an unbiased mask to guide image editing. The method comprises a Bias Elimination Cycle (BE-Cycle) that consists of forward and inverted paths, each incorporating a Structural Consistency Cycle (SC-Cycle) to maintain image content integrity during the editing process. Given a target text, the forward path utilizes a pre-trained model to synthesize the edited image. The inverted path then converts the edited image back to the source image using a source text that describes the original image. 

Additionally, the method allows for editing of both real images and synthetic images generated by other models, and the source text need not be the one used to generate the original image, so long as it adequately describes its main content. 
Specifically, the source text could be automatically produced by using an off-the-shelf captioning model~\cite{mokady2021clipcap} or obtained by making slight modifications, such as changing, adding, or removing several words from the target text. As shown in Fig~\ref{fig:model}, an example of this would be the removal of the phrase "and wearing a scarf.
The resulting processed source image and edited image conform to the same distribution, enabling the automatic generation of an unbiased mask through the identification of differences in their features. This is achieved through an improved mask computing pipeline that utilizes the encoder of CLIP~\cite{radford2021learning} to extract features from both images. The generated mask is then employed to guide unbiased semantic image editing. Comprehensive experiments demonstrate the proposed method's superiority, which significantly improves quantitative metrics and qualitative visual results.

\section{The Proposed Method}

\subsection{Preliminaries}
\label{sec2_1}
\textbf{Diffusion probabilistic models (DPMs).} 
DPMs~\cite{ho2020denoising,song2019generative,song2020score,watson2021learning} are a class of generative models that attempt to estimate the underlying data distribution $p(\bx)$ by gradually denoising a normally distributed variable. This can be seen as learning the inverse procedure of a fixed Markov Chain of length $T$.
In the context that semantic image editing is an extension of the text-to-image generation task, our current focus is solely towards text-conditioned DPMs.
During the inference stage, when given a text query $\bc$, DPMs utilize a parametric mean estimator $\mu_{\theta}$ to progressively denoise a synthesized image sampled from white Gaussian noise $\bx_T\sim{\mathcal{N}(\mathbf{0}, \bm{I})}$. 
Using the reparameterization trick, we can sample $\bx_{t-1}\sim{\mathcal{N}(\mu_{\theta}(\bx_t, \bc, t),\sigma_t^2\bm{I})}$ as: 
\begin{equation}
\bx_{t-1}= \mu_{\theta}(\bx_t, \bc, t)+\sigma_t\be_t,\, t=  1,\cdots,T,
\label{eq:DPM}
\end{equation}
where $\sigma_t$ is the standard deviation in the $t$-th step and $\be_t\sim\mathcal{N}(\mathbf{0},\bm{I})$. Note that the image generation direction proceeds from $\bx_T$ to $\bx_0$. To simplify notation, we use $\bz:=\bx_T\oplus\be_{1:T}$ as the latent code throughout this paper, where $\oplus$ denotes concatenation. Consequently, the mapping from $\bx_T$ to $\bx_0$ can be represented as $\bx_0=G(\bz,\bc)$.

\textbf{DPM-Encoder.} 
To obtain the latent code $\bz$, Wu et al.~\cite{wu2022unifying} proposed DPM-Encoder, an invertible encoder for stochastic DPMs. 
In detail, stochastic DPMs define a posterior distribution $q(\bx_{1:T}|\bx_0)$~\cite{ho2020denoising,song2020denoising} in the forward diffusion process. Then DPM-Encoder produces noisy images $\bx_1,\cdots,\bx_T$ from $q(\bx_{1:T}|\bx_0)$ and computes the $\be_t$. Formally, the overall sampling process of DPM-Encoder $\bz\sim{q_{\text{DPMEnc}}(\bz|\bx_0,\bc,\mu_{\theta})}$ can be defined as:
\begin{equation}
\begin{aligned}
    \bx_1&,\cdots,\bx_{T} \sim q(\bx_{1:T}|\bx_0), \\ 
    \be_t \,& = (\bx_{t-1}-\mu_{\theta}(\bx_t,\bc,t))/\sigma_t, \; t=1,\cdots,T, \\
    \bz:&=\bx_T\oplus\be_{1:T}.
\end{aligned}
\end{equation}

\subsection{Structural Consistency Cycle}
\label{sec2_2}
The goal of SC-Cycle~\cite{wu2022unifying} is to preserve the integrity of the image content during the editing process. 
The framework is shown in the top-right corner of Fig~\ref{fig:model}. Given a source image $\bx_0^{\text{source}}$ and a source description $\bc^{\text{source}}$, SC-Cycle first encodes the latent code $\bz^{\text{source}}$ by a DPM-Encoder. 
Then the edited image $\bx_0^{\text{target}}$ is generated by applying the mapping $G$ with the target text $\bc^{\text{target}}$:
\begin{equation}
\begin{split}
    & \bz^{\text{source}} \sim\,  q_{\text{DPMEnc}}(\bz|\bx_0^{\text{source}},\bc^{\text{source}},\mu_{\theta}),\\
    & \bx_0^{\text{target}}\,=\, G(\bz^{\text{source}},\bc^{\text{target}})=G(\bx_T^{\text{source}}\oplus\be_{1:T}^{\text{source}},\bc^{\text{target}})
    .
\end{split}
\label{eq:sc-cycle}
\end{equation}

The edited image $\bx_0^{\text{target}}$ synthesized by SC-Cycle~\cite{wu2022unifying} preserves the details specific to the source image $\bx_0^{\text{source}}$ by utilizing the latent code $\bz^{\text{source}}$. This property leads to the similarity in structure between $\bx_0^{\text{source}}$ and $\bx_0^{\text{target}}$.

\subsection{Bias Elimination Cycle}
\label{sec2_3}
BE-Cycle aims to automatically generate an unbiased mask by contrasting the generated images of the forward path and the inverted path. As illustrated in the top-left corner of Fig~\ref{fig:model}, each path comprises an SC-Cycle. In the forward path, $\bx_0^{\text{target}}$ is synthesized following Eq.~\eqref{eq:sc-cycle}. Then in the inverted path, we attempt to restore the source image from $\bx_0^{\text{target}}$, conditioned on the source text $\bc^{\text{source}}$:
\begin{equation}
\begin{split}
    &\bz^{\text{target}}\sim\, q_{\text{DPMEnc}}(\bz|\bx_0^{\text{target}},\bc^{\text{target}},\mu_{\theta}),\\
    &\bx_0^{\text{inv}}\,\,\,\,=\, G(\bz^{\text{target}},\bc^{\text{source}})=G(\bx_T^{\text{target}}\oplus\be_{1:T}^{\text{target}},\bc^{\text{source}}).
\end{split}
\end{equation}
The forward and inverted paths of BE-cycle both rely on the same pre-trained model, resulting in a similar contextual prior bias in the generated images $\bx_0^{\text{target}}$ and $\bx_0^{\text{inv}}$. In other words, both images are conforming to the same distribution. Therefore, the contextual prior bias does not affect identifying the regions that need to be edited by contrasting $\bx_0^{\text{target}}$ and $\bx_0^{\text{inv}}$.

To produce the mask, the first step is to extract the visual features of $\bx_0^{\text{target}}$ and $\bx_0^{\text{inv}}$ using a visual encoder $E_{\text{CLIP}}$ from a pre-trained CLIP model~\cite{radford2021learning}:
\begin{equation}
    \bF=E_{\text{CLIP}}(\bx_0),\; \bF\in\mathbb{R}^{m\times n\times d},
\end{equation}
where $\mathbb{R}^{m\times n}$ is the spatial space, $d$ is the feature dimension, and $\bF_{i,j,:}$ is the image grid feature with the grid's coordinate $(i,j)$, for $i\in\{1,2,\cdots,m\}$ and $j\in\{1,2,\cdots,n\}$. 
To measure the degree of change in each region, the cosine similarity between $\bF_{i,j,:}^{\text{target}}$ and $\bF_{i,j,:}^{\text{inv}}$ is calculated by:
\begin{equation}
    \bS_{i,j} = \cos{\left\langle\bF_{i,j,:}^{\text{target}},\bF_{i,j,:}^{\text{inv}}\right\rangle}.
\end{equation}
Finally, we can obtain the mask $\bM$ by:
\begin{equation}
    \bM_{i,j}=
    \begin{cases}
        1, & \mbox{if } \frac{{\rm abs}(\bS_{i,j})-\min({\rm abs}(\bS))}{\max({\rm abs}(\bS))-\min({\rm abs}(\bS))} > \delta,\\
        0, & \mbox{else},
    \end{cases}
\end{equation}
where $\delta$ is a threshold to control the binarization of the mask, $\text{max}(\cdot)$, $\text{min}(\cdot)$ obtain the maximal value, and the minimal value of a matrix, respectively, while $\text{abs}(\cdot)$ returns the absolute value of each element in a matrix or a scalar input. In practice, we set $\delta=0.5$. To increase the resolution of the mask, the image is divided into $2\times 2$ grids, each of which respectively generates a mask. 
All grid masks are then assembled to get the final mask according to their spatial positions. Finally, the mask is resized to match the spatial size of $\bx_0$.

\subsection{Mask-Guided Unbiased Editing}
\label{sec2_4}
In the final stage, the generated unbiased mask is used to guide the image editing process. The pipeline is illustrated in the left-bottom corner of Fig~\ref{fig:model}. 
As pointed in \cite{balaji2022ediffi}, the text-conditioned DPMs mainly rely on the text prompt to guide the sampling process at the early sampling stage, while gradually shifting towards visual features, as the generation continues.
Therefore, we utilize the text prompt to guide the generation within the mask, while keeping the regions outside the mask as similar to the source image as possible. For convenience, we divide the mapping $G$ into two steps: $G(\cdot)=G_2(G_1(\cdot))$. In the first step, we obtain the image strongly influenced by the target text:
\begin{equation}
    \bx_k^{\text{target}}=G_1(\bx_T^{\text{source}}\oplus\be_{(k-1):T}^{\text{source}},\bc^{\text{target}}),
\end{equation}
where $k\in\{1,2,\cdots,T\}$ indicates the $k$-th step of the diffusion process. Then we sample $\bx_k^{\text{source}}$ from the posterior distribution $q(\bx_k|\bx_0^{\text{source}})$ and optimize the image by the mask:
\begin{equation}
    \bx_k^{\text{opt}}=\bM\odot\bx_k^{\text{target}}+(\mathbf{1}-\bM)\odot\bx_k^{\text{source}},
\end{equation}
where $\odot$ is the element-wise product.
Finally, the edited image without the effect of contextual prior bias is synthesized by:
\begin{equation}
    \bx_0^{\text{opt}}=G_2({\bx_k^{\text{opt}}\oplus\be_{1:k}^{\text{source}},\bc^{\text{target}}}).
\end{equation}
The unbiased mask only focuses on the regions that need editing, thus avoiding unexpected modification on the final edited image $\bx_0^{\text{opt}}$, such as the cat ears.

\section{Experiments}

\begin{table}[]
\vspace{-0 mm}
\caption{Quantitative results on the zero-shot semantic image editing dataset~\cite{wu2022unifying}.
$*$ denotes that we ran only 1 trial for each hyperparameter combination, while we ran 15 trials in the rest experiments.
\colorbox{lightyellow}{Best} and \colorbox{lightgray}{Second} results are in highlight.
}  
\label{tab_SoTA}
\vspace{-1mm}
\centering
\resizebox{0.9\linewidth}{!}{
\begin{tabular}{llcccc}
\toprule
         & \textbf{Method}         & $\mathcal{S}_{\textbf{CLIP}}\uparrow$           & $\mathcal{S}_{\textbf{D-CLIP}}\uparrow$         & \textbf{PSNR}$\uparrow$           & \textbf{SSIM}$\uparrow$           \\ \midrule
\multicolumn{1}{c}{\multirow{4}{*}{LDM-400M}}          & SDEdit~\cite{meng2021sdedit}         & 0.332          & 0.264          & 13.68          & 0.390          \\
         & DiffEdit~\cite{couairon2022diffedit} & 0.323      & 0.208            & 18.50       & 0.588 \\ 
         & CycleDiffusion~\cite{wu2022unifying} & \bestcell{0.333}          & \secondcell{0.275}          & \secondcell{18.72}          & \secondcell{0.625}          \\
         & \textbf{Ours}           & \bestcell{0.333} & \bestcell{0.281} & \bestcell{19.12} & \bestcell{0.635} \\ \midrule
\multicolumn{1}{c}{\multirow{4}{*}{SD-v1-1}}          & SDEdit~\cite{meng2021sdedit}         & \bestcell{0.339} & 0.248          & 15.50          & 0.498          \\
       & DiffEdit~\cite{couairon2022diffedit} & 0.316   
                & 0.175        &\bestcell{22.47}  & 0.729 \\
         & CycleDiffusion~\cite{wu2022unifying} & 0.331          & \secondcell{0.262}          & 21.98          & \secondcell{0.731}          \\
         & \textbf{Ours}           & \secondcell{0.332}          & \bestcell{0.265} & \secondcell{22.21} & \bestcell{0.734} \\ \midrule
\multicolumn{1}{c}{\multirow{4}{*}{SD-v1-4 }}   & SDEdit~\cite{meng2021sdedit}         & \bestcell{0.344} & 0.258          & 15.93          & 0.512          \\
       & DiffEdit~\cite{couairon2022diffedit} & 0.318      & 0.172          & \bestcell{22.70}   & 0.731 \\
         & CycleDiffusion~\cite{wu2022unifying} & 0.334          & \secondcell{0.272}          & 21.92          & \bestcell{0.731} \\
         & \textbf{Ours}           & \secondcell{0.337}          & \bestcell{0.283} & \secondcell{22.08} & \secondcell{0.730}          \\ \midrule
         \multicolumn{1}{c}{\multirow{4}{*}{SD-v1-4 }}   & SDEdit*~\cite{meng2021sdedit}         & \bestcell{0.335} & \secondcell{0.221}          & 15.64          & 0.505          \\
       & DiffEdit*~\cite{couairon2022diffedit} & 0.305      & 0.088          & \bestcell{22.72}   & \secondcell{0.733} \\
         & CycleDiffusion*~\cite{wu2022unifying} & 0.327          & 0.206          & 21.85          & 0.721 \\
         & \textbf{Ours*}           & \secondcell{0.331}          & \bestcell{0.235} & \secondcell{22.71} & \bestcell{0.741}          \\  \bottomrule
\end{tabular}
}
\vspace{-1 mm}
\end{table}

\begin{table}[]
\vspace{-1 mm}
\caption{Comparisons by using different masks with/without the CLIP~\cite{radford2021learning} encoder. The biased/unbiased mask is generated by contrasting the input and output of the forward/inverted path in the BE-Cycle. All the experiments are conducted with the SD-v1-4 pre-trained model.
}  
\label{tab_ablation}
\vspace{-1mm}
\centering
\resizebox{0.9\linewidth}{!}{
\begin{tabular}{cccccc}
\toprule
              &  w/ CLIP Encoder        & $\mathcal{S}_{\textbf{CLIP}}\uparrow$           & $\mathcal{S}_{\textbf{D-CLIP}}\uparrow$         & \textbf{PSNR}$\uparrow$           & \textbf{SSIM}$\uparrow$           \\ \midrule
\multicolumn{1}{c}{\multirow{2}{*}{w/ Biased Mask}}   & \XSolidBrush               & 0.336 & 0.281 & 21.49 & 0.725 \\
                                  & \CheckmarkBold              & \bestcell{0.338}            & 0.281                     & 21.85                     & 0.724                     \\ \midrule
\multicolumn{1}{c}{\multirow{2}{*}{w/ Unbiased Mask}} & \XSolidBrush               & 0.337 & 0.282 & 21.90 & 0.725 \\
                                  & \CheckmarkBold               & 0.337                     & \bestcell{0.283}            & \bestcell{22.08}            & \bestcell{0.730}            \\ \bottomrule
\end{tabular}
}
\vspace{-5 mm}
\end{table}

\subsection{Implementation Details}\label{sec3_1}
Experiments were conducted on the zero-shot semantic image editing dataset~\cite{wu2022unifying}, which was specifically gathered for semantic image editing task. The dataset consists of a set of 150 tuples, each containing a source image, a source text, and a target text. For a fair comparison, DiffEdit~\cite{couairon2022diffedit} used the settings from its original paper and the remaining experiments followed the settings of CycleDiffusion~\cite{wu2022unifying} to use the DDIM sampler ($\eta=0.1$) with 100 steps, set the classifier-free guidance scale of the encoding process as 1, and enumerate the classifier-free guidance scale of the decoding step as $\{1, 1.5, 2, 3, 4, 5\}$. 
To preserve image content, editing began with an image that was not fully noised. 
Therefore, we enumerated the step of adding noise as $\{85, 80, 75, 70, 60, 50\}$.\footnote{It is the same with the early stop step of $\{15, 20, 25, 30, 40, 50\}$ in \cite{wu2022unifying}.} Then the optimization step $k$ was selected from $\{85, 80, 75, 70, 60, 50\}$.
We ran 15 trials for each hyperparameter combination. To remove the effect of random noise in the mask computing process, we averaged all the masks generated with different hyperparameter combinations. The final results were automatically selected based on the directional CLIP score $\mathcal{S}_{\textbf{D-CLIP}}$. 

\textbf{Metrics:} To evaluate editing performance over all comparison methods and our Dual-Cycle Diffusion, we adopted four metrics to evaluate edited images' faithfulness to source images and authenticity to target texts. They are {\textbf{SSIM}}, {\textbf{PSNR}}, CLIP score $\mathcal{S}_{\textbf{CLIP}}$~\cite{radford2021learning}, and directional CLIP score $\mathcal{S}_{\textbf{D-CLIP}}$~\cite{patashnik2021styleclip}. {\textbf{SSIM}} is used to measure the similarity between two images, and {\textbf{PSNR}} is used to quantify image quality. $\mathcal{S}_{\textbf{CLIP}}$ can assess the alignment between the generated image and the target text, while $\mathcal{S}_{\textbf{D-CLIP}}$ can evaluate the similarity between the images' and texts' changes.
In addition, we conducted a user study to further evaluate the effectiveness of the model in eliminating contextual prior bias. The participants were instructed to vote for the images that had fewer unexpected modifications according to the target texts and source images, resulting in $1000$ votes per method-to-method comparison. The results are shown in Table~\ref{tab_user_study}.

\subsection{Comparisons with Existing Methods}\label{sec3_2}
To validate the effectiveness of the proposed Dual-Cycle Diffusion, we compared our model with other methods based on diffusion model, including SDEdit~\cite{meng2021sdedit}, DiffEdit~\cite{couairon2022diffedit}, and CycleDiffusion~\cite{wu2022unifying}. To investigate the influence of data size, data quality, and training details on models' performance, several pre-trained text-to-image diffusion models are used: (1) LDM-400M, an LDM model~\cite{rombach2022high} with 1.45B parameters, trained on LAION-400M~\cite{schuhmann2021laion}, (2) SD-v1-1, a Stable Diffusion model~\cite{rombach2022high} with 0.98B parameters, trained on LAION-5B~\cite{schuhmann2022laion}, (3) SD-v1-4, finetuned from SD-v1-1 for improved aesthetics and classifier-free guidance sampling.

The quantitative results by the involved competitors are presented in Table~\ref{tab_SoTA}. The proposed method achieved the best or comparable scores on all metrics among the three pre-trained models, indicating its ability to enhance both edited images' faithfulness to source images and authenticity to target texts. Comparisons of the results of LDM-400M and SD-v1-4 suggest that large training data size can aid in comprehending image contents, leading to better preservation of image content when using SD-v1-4. Additionally, we can observe from a comparison of the SD-v1-1 and SD-v1-4 results that the improved classifier-free guidance sampling is useful for synthesizing edited images that are more authentic to target texts. Overall, using SD-v1-4 as the pre-trained model resulted in the best performance. Hence, all subsequent experiments and comparisons will be based on the SD-v1-4 model.

\begin{figure}[!t]
	\centering \vspace{-0 mm}
	\includegraphics[width=\linewidth]{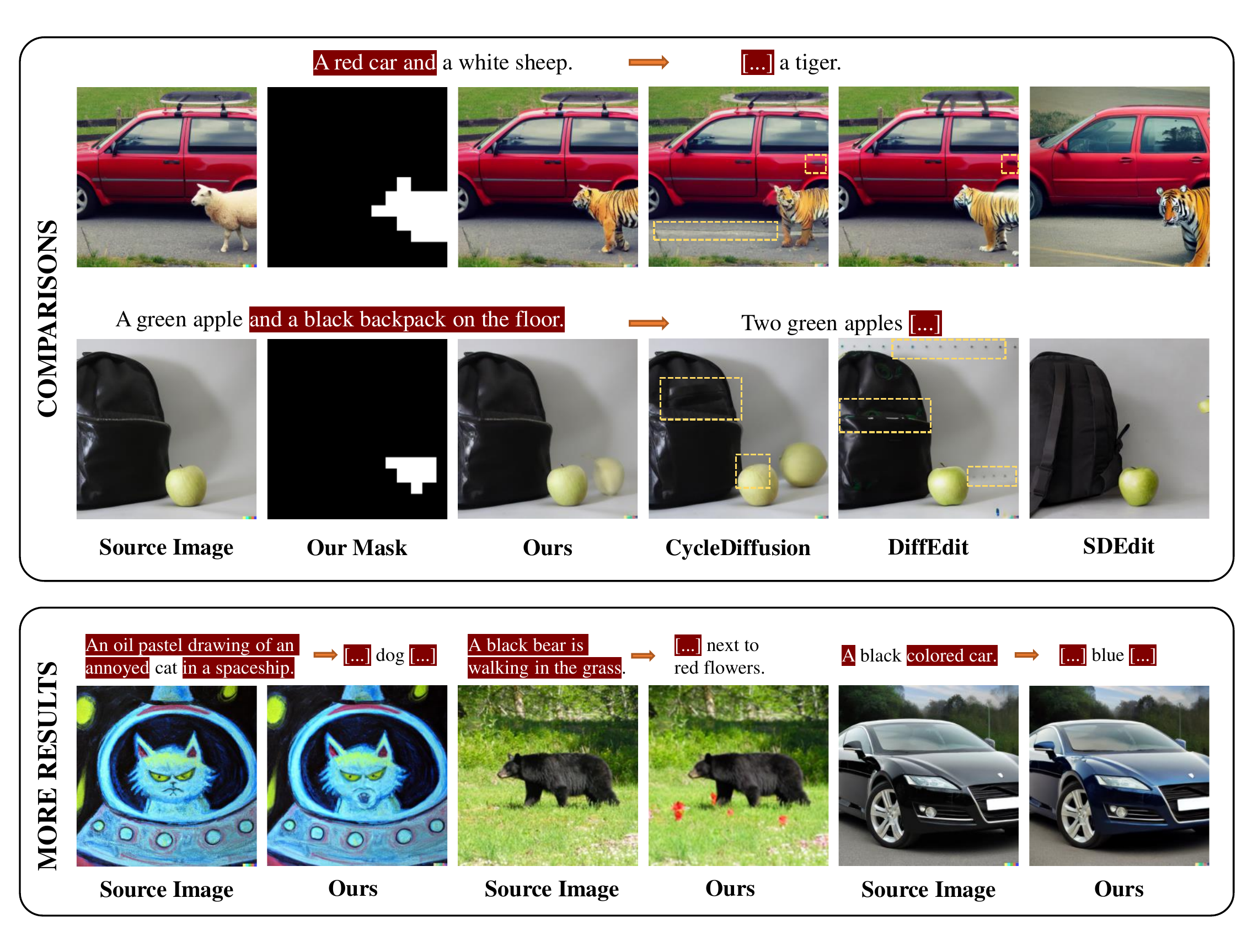}
    \vspace{-8 mm}
	\caption{The masks and edited samples generated by our proposed method. The first two rows show comparisons with two most representative baseline models: SDEdit~\cite{meng2021sdedit}, DiffEdit~\cite{couairon2022diffedit}, and CycleDiffusion~\cite{wu2022unifying}. Among all samples, our method outperforms the other models in image content preservation. Yellow boxes mark unexpected modifications, and the masks reveal the specific edited areas. The last row shows more results generated by the proposed method.}
 	\vspace{-4 mm}
	\label{fig:sota}
\end{figure}

In Fig~\ref{fig:sota}, we provide several visual comparisons of edited results generated by different methods.
According to the visual comparisons, we can observe that the proposed method not only has accomplished modifications according to the text, but also has better image content preservation. For instance, as shown in the first row of Fig~\ref{fig:sota}, the goal is to replace the sheep with a tiger. SDEdit made modifications to the entire image, while both DiffEdit and CycleDiffusion were able to maintain most parts of the source image. But they still unexpectedly added a door handle to the car, which was marked by a yellow box. This observation can also demonstrate the existence of contextual prior bias. 
Furthermore, the second row aims to add an apple. CycleDiffusion failed to preserve the shapes of the backpack and the apple already in the source image, while our method succeeded. DiffEdit even failed to produce another apple. The third row provides more results generated by the proposed method.

\begin{table}[]
\vspace{-1 mm}
\caption{User study results. The participants were instructed to vote for the images that had fewer unexpected modifications according to the target texts and source images.
}  
\label{tab_user_study}
\vspace{-1mm}
\centering
\resizebox{0.85\linewidth}{!}{
\begin{tabular}{cccc}
\toprule
              &  CycleDiffusion        & Ours w/ Biased Mask           & Ours w/ Unbiased Mask         \\ \midrule
Votes (\%)   & 42.7               & 46.8 & \textit{Reference}                     \\  \bottomrule
\end{tabular}
}
\vspace{-4.5 mm}
\end{table}

\subsection{Ablation Study}\label{sec3_3}

\begin{figure}[!t]
	\centering \vspace{-0 mm}
	\includegraphics[width=0.9\linewidth]{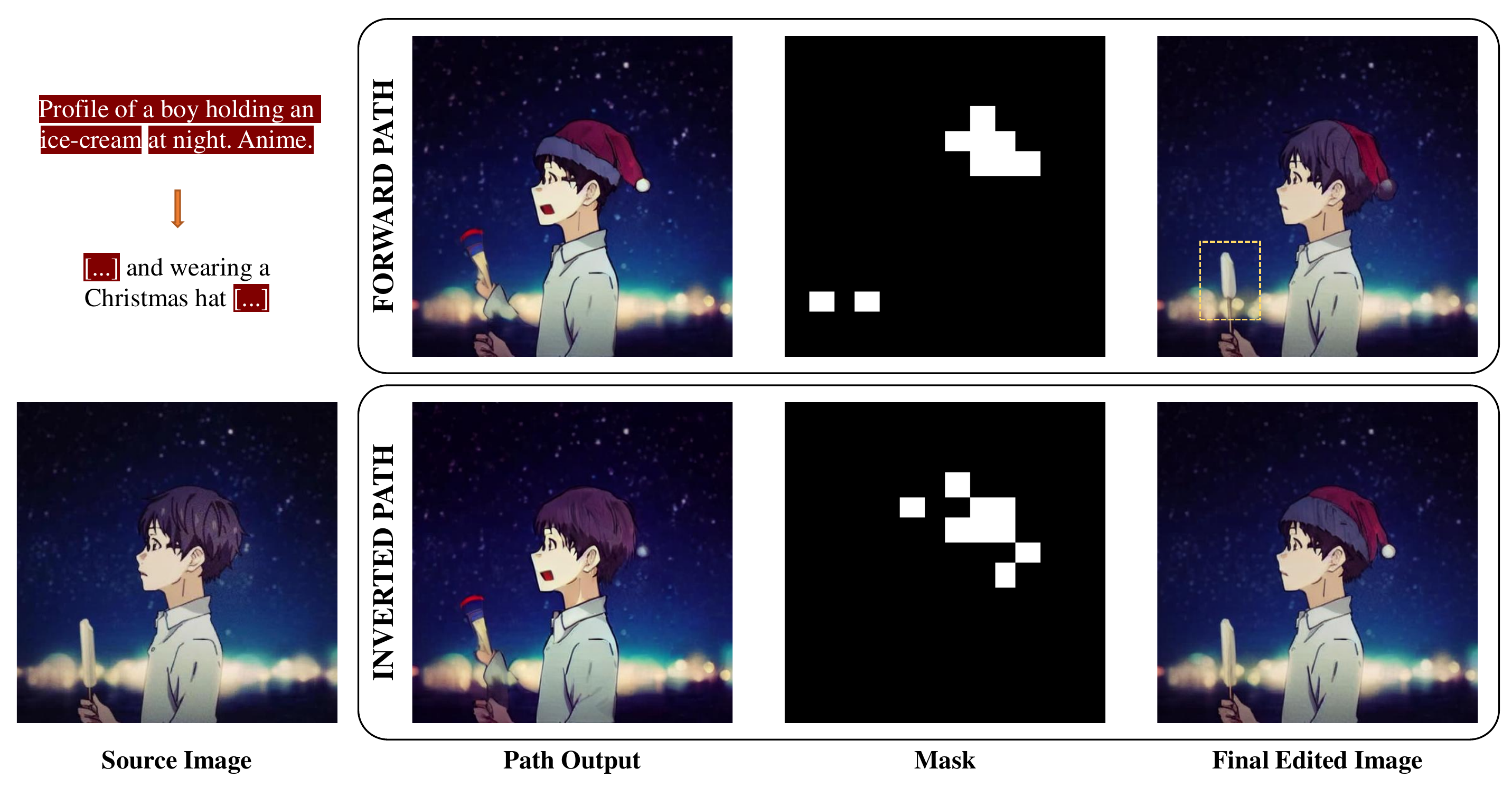}
    \vspace{-2 mm}
    \caption{The BE-Cycle's forward path and inverted path's masks, outputs, and final edited results. Specifically, the mask is generated by comparing the input and output of the forward or inverted path. To remove the effect of random noise in the mask computing process, we averaged all the masks generated with different hyperparameter combinations. From the comparisons, we can obtain an ice-cream similar to the source image with the unbiased mask.}
 	\vspace{-5.5 mm}
	\label{fig:ablation}
\end{figure}

In this section, we aim to first explore the effectiveness of the BE-Cycle and the improved mask computing pipeline, and then further demonstrate the existence of contextual prior bias. 
To this end, we conducted additional experiments based on the biased mask generated by contrasting $\bx_0^{\text{source}}$ and $\bx_0^{\text{target}}$ with the same process to calculate the unbiased mask.
Additionally, we exhibited the results without the CLIP encoder. Here, we used SD-v1-4 as the pre-trained model. The quantitative results are reported in Table~\ref{tab_ablation}.
From the results, we can observe that all the models obtained similar $\mathcal{S}_{\textbf{CLIP}}$ and $\mathcal{S}_{\textbf{D-CLIP}}$ scores. However, the models, using the CLIP encoder, achieved better image content preservation, which demonstrates the effectiveness of the proposed mask computing pipeline.  
Then, we discuss the effects of the mask by comparing the models using different masks. As seen from the results with the CLIP encoder, the model using the unbiased mask produced from the inverted path achieved a significant improvement in the {\textbf{SSIM}} and {\textbf{PSNR}} scores. This means that the inverted path increases the accuracy of identifying the regions that need to be edited. In other words, we can achieve better performance with fewer modifications. The reason why using the biased mask resulted in worse performance is the existence of contextual prior bias, which causes unexpected modifications. In short, the BE-Cycle is effective in eliminating the contextual prior bias derived from the pre-trained text-to-image models.

Additionally, we also depict the visual comparisons of the masks and outputs of the BE-Cycle's paths in Fig~\ref{fig:ablation}. The final edited images using different masks are also provided. 
Here, we averaged all the masks generated with different hyperparameter combinations to remove the noise's effect (\textit{i.e.}, the boy's open mouth) in the mask computing process. 
From the results of the forward path, it can be observed that using a pre-trained model changed some image contents, such as the ice-cream's shape. This change is also reflected in the mask. Therefore, using the biased mask, the model shortened the ice-cream's length in the final edited image. In contrast, the unbiased mask produced from the inverted path can focus on the regions of the Christmas hat, which leads to synthesizing an ice-cream as similar to the source image as possible. In a word, the proposed method can achieve the goal of editing the image without the contextual prior bias's influence.

\section{Conclusion}

In this paper, we address the issue of contextual prior bias in semantic image editing and propose Dual-Cycle Diffusion to eliminate its effects by generating an unbiased mask to guide image editing. 
Our method employs SC-Cycle and BE-Cycle to eliminate distributional shifts between source and generated images from a pre-trained model. 
Leveraging CLIP's encoder to extract image visual contents, our model could produce an unbiased mask by identifying content differences in each region. 
This allows our model to focus on the regions that need editing without the effects of contextual prior bias.
Through extensive experiments and ablation studies, we validate the superiority of Dual-Cycle Diffusion over other diffusion-based semantic image editing methods.

\normalem
\bibliographystyle{ieee_fullname}
\bibliography{Ref/citations}


\end{document}